%% file: FbGLMM.tex
\documentclass[a4paper,american,reqno]{amsart}

\input{setup-preprint}
\bibliography{FbGLMM}

\begin{document}

\title[Fair Generalized Linear Mixed Models]%
{Fair Generalized Linear Mixed Models}

\author[J. P. Burgard, J. V. Pamplona]%
{Jan Pablo Burgard, João Vitor Pamplona}

\address[J. P. Burgard, J. V. Pamplona]{%
  Trier University,
  Department of Economic and Social Statistics,
  Universitätsring 15,
  54296 Trier,
  Germany}
\email{burgardj@uni-trier.de, pamplona@uni-trier.de}

\date{\today}

\begin{abstract}
When using machine learning for automated classification, it is important to account for fairness in the classification. 
Fairness in machine learning aims to ensure that biases in the data and model inaccuracies do not lead to discriminatory decisions. 
E.g., classifications from fair machine learning models should not discriminate against sensitive variables such as sexual orientation and ethnicity. 
The training data often in obtained from social surveys. 
In social surveys, oftentimes the data collection process is a strata sampling, e.g. due to cost restrictions.
In strata samples, the assumption of independence between the observation is not fulfilled. 
Hence, if the machine learning models do not account for the strata correlations, the results may be biased. 
Especially high is the bias in cases where the strata assignment is correlated to the variable of interest. 
We present in this paper an algorithm that can handle both problems simultaneously, and we demonstrate the impact of stratified sampling on the quality of fair machine learning classifications in a reproducible simulation study.
\end{abstract}

\keywords{\input{keywords}}
\subjclass[2020]{\input{msc2020}}

\maketitle

\input{introduction}

\input{chapter1}

\input{chapter4}
\input{chapter5}

\input{conclusion}

\input{acknowledgements}

\printbibliography

\appendix
\input{appendix}

\end{document}

%% file: setup-preprint.tex
\usepackage[utf8]{inputenc}
\usepackage[T1]{fontenc}
\usepackage{babel}
\usepackage[binary-units=true]{siunitx}
\usepackage[ruled,linesnumbered]{algorithm2e}
\usepackage{csquotes}
\usepackage{todonotes}
\usepackage{multicol}
\usepackage{comment}
\usepackage{booktabs}
\usepackage{flafter}
\usepackage{longtable}
\usepackage{dsfont}
\usepackage{placeins}
\usepackage{float}
\allowdisplaybreaks
\usepackage[style = authoryear-comp,
            maxbibnames = 100,
            maxcitenames = 2,
            giveninits = true,
            uniquename = init,
            isbn = false,
            backend = bibtex]{biblatex}
\usepackage[colorlinks,
            citecolor=blue,
            urlcolor=blue,
            linkcolor=blue]{hyperref} 

\makeatletter
\patchcmd{\@settitle}{\uppercasenonmath\@title}{\scshape\large}{}{}
\patchcmd{\@setauthors}{\MakeUppercase}{\scshape\normalsize}{}{}
\makeatother

\vfuzz \hfuzz
\makeatletter
\@namedef{subjclassname@2020}{%
  \textup{2020} Mathematics Subject Classification}
\makeatother

\input{macros}


%% file: macros.tex
\SetKwInOut{Input}{Input}
\SetKwInOut{Output}{Output}

\newcommand{\define}{\mathrel{{\mathop:}{=}}}



\newcommand\MyBox[2]{
  \fbox{\lower0.75cm
    \vbox to 1.7cm{\vfil
      \hbox to 1.7cm{\hfil\parbox{1.4cm}{#1\\#2}\hfil}
      \vfil}%
  }%
}



%% file: keywords.tex
Logistic Regression, 
Fair Machine Learning,
Mixed Models.

%% file: msc2020.tex
90C90,
90-08,
68T99
%
%

%% file: introduction.tex
\section{Introduction}
\label{sec:introduction}

With the advent of automatic decision-making, the need for fair decision-making algorithms is steadily rising. The automatic decision should comply to restriction based on societal values, such as non-discrimination of parts of the population. Machine learning algorithms, while offering efficiency, can inadvertently perpetuate bias in critical areas like loan approvals \parencite{das2021fairness} and criminal justice \parencite{green2018fair}. In loan applications, factors like marital status can lead to unfair disadvantages for single individuals, while in criminal justice, algorithms might associate race with recidivism risk, leading to discriminatory sentencing despite individual circumstances. This highlights the need for fair and unbiased AI frameworks to ensure equal opportunities and outcomes for all.

The training data, used to learn the machines for the automatic decision-making, oftentimes comes from surveys. These surveys typically are drawn according to a sampling plan, and hence do not comply with the general assumption in machine learning, that each unit is sampled independently and with the same probability of inclusion. For a detailed discussion on survey methods including sampling strategies, see \textcite{Lohr2009}.

The field of fair machine learning has thrived, with numerous research articles exploring approaches to mitigate bias in various algorithms. Notable examples include fair versions of Logistic and Linear Regression \parencite{fairRL2}, Support Vector Machines \parencite{FairSVM2}, Random Forests \parencite{fairRF}, Decision Trees \parencite{FairDT}, and Generalized Linear Models (GLMs) \parencite{fairGLM}. These methods aim to address potential discrimination arising from historical data or algorithmic design, ensuring fairer outcomes for all individuals.

In this paper we propose a Generalized Mixed Model for fair classifications. We show how to estimate the model and evaluate its performance against the current model that does not take the possible clustering of the data into account. As far as we know, this has not been proposed before.

The paper is organized as follows: In Section \ref{sec:chapter-1} we establish the theoretical underpinnings of fair generalized linear mixed models and propose a strategy for solving them.
In Section \ref{sec:chapter-3}, we conduct a comprehensive evaluation of our proposed method's effectiveness through various tests.
Finally, in Section \ref{sec:chapter-5}, we demonstrate the practical applicability of our algorithm by solving a real-world problem using the Bank marketing dataset \parencite{misc_bank_marketing_222}.
Our key findings and potential future directions are presented in Section \ref{sec:conclusion}.

%% file: chapter1.tex
\section{Fair Generalized Linear Mixed Models}\label{sec:chapter-1}

In recent years, there has been a growing interest in developing fair machine learning algorithms. Fairness is a complex concept, but it generally refers to the idea that algorithms should not discriminate against certain groups of people \parencite{importance}. This is important because algorithms are increasingly being used to make decisions about people's lives, such as whether to grant them a loan or admit them to college. 

In this context, we have Generalized Linear Models (GLMs) that are a class of models that can be used to model a variety of response variables, including count, continuous and binary data that is the focus of this work. Following the same idea, but with some changes, we have the Generalized Linear Mixed Models that allow for the inclusion of random effects, which are random variables that capture the variability in the response variable due to the hierarchical data structure. GLMMs are a powerful tool for analyzing data that are non-normal and hierarchical. They are used in a wide variety of fields, including medicine and psychology \parencite{psicologia,medicina} for example. However, GLMMs, like many other statistical models, can lead to unfair outcomes.

\subsection{Fair Classifications}

In classification algorithms, we need to find a function that predicts the label $y$ given a feature vector $x \in \mathbb{R}^p$. This function is learned on a training set $D = {(x^{\ell}, y_{\ell})}^N_{\ell=1}$. One typically way to do that is minimizing a loss function $L(\beta)$ over a training set that minimize the classification error in this set.

In the context of fairness in binary classification, each observation $\ell$ has an associated sensitive feature $s_\ell \in \{0, 1\}$, and the objective then becomes finding a solution with good accuracy (AC) while it is also fair. The concept of fairness in machine learning can be seen from different metrics, here, we study the concept of disparate impact (DI) \parencite{bigdata}. Other unfairness metrics for binary classifiers can be found in \textcite{Paper1}.

Disparate impact refers to a situation where the classification of a model disproportionately harm points with different sensitive feature values. That is, a classifier is considered fair with respect to disparate impact if the probability, of the model $(\mathbb{P}_m)$, of its prediction remains constant across both values of the sensitive feature $s$, i.e.,
\begin{equation*}
\mathbb{P}_m(\hat{y}_\ell = 1 | s_\ell=0, X) = \mathbb{P}_m(\hat{y}_\ell = 1 | s_\ell=1, X).
\end{equation*}

We can also say that an algorithm suffers from disparate impact if the decision-making process grants a disproportionately large fraction of beneficial outcomes to certain sensitive feature groups.

\subsection{Generalized linear mixed model}
Generalized linear mixed models are regarded as an extension of generalized linear models that effectively incorporate random effects. These random effects can come from a survey that has strata bias, for example. This section provides a brief explanation of how we can model the GLMMs as can be seen in \textcite{stroup2012generalized}.

Let $y_{ij}$ denote the label in the observation $j$ in strata $i$, where $i ~\in~ [1,n] ~:=~ \{1, \dots, n\}$ and $j~\in~[1, T_i]$, with $T_i$ being the size of the strata $i$. These observations are collected in the vector $y^\top_i~=~(y_{i_1}, \dots, y_{i_{T_{i}}})$. Let $(x^{ij})^\top~=~(x^{ij}_1, \dots, x^{ij}_p)$ represent the covariate vector associated with fixed effects, and $(z^{ij})^\top = (z^{ij}_1, \dots, z^{ij}_n) \in \mathbb{R}^{n}$ denote the covariate vector associated with the random effects $b_i ~\in~ [1,n]$ that follow a normal distribution with a covariance matrix $Q_b = Blockdiag(Q, \dots, Q) \in \mathbb{R}^{n \times n}$.

The generalized linear mixed model can be expressed as follows:
\begin{equation}\label{eq: 1}
g(\mu_{ij}) = \beta_0 + \beta^\top x ^{ij} + (z^{ij})^\top b_i.
\end{equation}
Here, $g$ represents a monotonic and continuously differentiable link function, $\mu_{ij} = E(y_{ij}|b_i,x^{i^j},z^{i^j})$, $\beta_0$ is the intercept, $\beta$ the fixed effects and $b_i$ represents the strata-specific random effects.

We can represent Model \eqref{eq: 1} using matrix notation. Let $(X^i)^\top~=~[x^{i_1}, \dots, x^{i_{T_i}}]\in\mathbb{R}^{p \times T_i}$ denote the design matrix for the $i$-th strata, and $\tilde{\beta}^\top  = (\beta_0, \beta^\top )$ represent the linear parameter vector, including the intercept. Let $\tilde{X}^i = [ \mathds{1}, X^i] \in \mathbb{R}^{(p+1)}$ be the corresponding matrix, where $\mathds{1}^\top  = (1, \dots, 1) \in \mathbb{R}^{T_i}$. By grouping the observations within each strata, the model can be represented as:
\[
g(\mu_i) = \tilde{X}^i \tilde{\beta} + Z^ib_i,
\]
where $(Z^i)^\top  = [z^{i_1}, \dots, z^{i_{T_i}}] \in \mathbb{R}^{n \times T_i}$. For all observations one obtains
\[
g(\mu) = \tilde{X}\tilde{\beta} + Zb,
\]
with $\tilde{X}  = [\tilde{X}^1 , \dots, \tilde{X}^n ] \in \mathbb{R}^{N \times (p+1)}$ and a block-diagonal matrix $Z~=~[Z^1, \dots, Z^n] \in \mathbb{R}^{N \times n}$, considering, w.l.o.g., that the first $T_1$ points, of $\tilde{X}$ belong to strata $1$, the next $T_2$ points belong to strata $2$, and so on, that is, there is an ordering, by strata, in the data.

For $r \in[1,p]$ and $i  \in[1,n]$, let us introduce the notation $(x^i)^\top_{r} ~=~ (x^{i_1}_r, \dots, x^{i_{T_{i}}}_r) \in \mathbb{R}^{T_i}$ to represent the covariate vector of the $r$-th fixed effect in strata $i$. Furthermore, we define $x^\top _{r} = ((x^1)^\top_{r}, \dots, (x^n)^\top_{r}) \in \mathbb{R}^{N}$. Consequently, the $r$-th design matrix, which includes the intercept and solely the $r$-th covariate vector, can be expressed as:

\[
X^i_{r} = [\mathds{1}, x^i_{r}] \in \mathbb{R}^{T_i \times 2}
\]
and
\[
X_{r} = [\tilde{\mathds{1}}, x_{r}] \in \mathbb{R}^{N \times 2},
\]
with
\[
\tilde{\mathds{1}} = (1, \dots, 1) \in \mathbb{R}^{N}
\]
representing the design matrices for stratas $i$ and the entire sample, respectively. Within strata $i$, the predictor that exclusively contains the $r$-th covariate takes the form of $\eta^i_{r} = X^i_{r} \tilde{\beta}_r + Z^ib_i$, where $\tilde{\beta}^\top _r = (\beta_0, \beta_r) \in \mathbb{R}^{2}$. For the entire sample, we obtain:

\[
\eta_{r} = X_{r} \tilde{\beta}_r + Zb
\]
and
\[
\eta^{i} = X^i \beta + Z^ib_i.
\]

Ignoring the mixed effects we can state that the logistics regression is a special case of GLMs. In the next chapters, we will see this case and some of its particularities for the case in which we have unfair datasets.

\subsection{Fair Logistic Regression}\label{sec2.2}
In classification algorithms that employ logistic regression that can be seen in \textcite{john2004mp}, a probabilistic model is used to link a feature vector $x$ to the class labels $y \in \{0,1\}$. The link function is:

\begin{equation*}
     p(\hat{y}=1|x,\beta) = m_{\beta}(x) =  \dfrac{1}{1 + e^{-\beta^\top x}},
\end{equation*}
where $\beta$ is obtained by solving the maximum likelihood problem on the training set ($D$), that is, $\beta^* = argmax_{\beta} \sum_{(x,y) \in D} \log p(y|x, \beta)$. Therefore, as we are working with a minimization problem, the corresponding loss function is defined as $- \sum_{(x,y) \in D} \log p(y|x, \beta)$, and the complete optimization problem is formulated as follows:

\begin{align}
\displaystyle \min_{\beta}\hspace{0.1cm} & -\displaystyle\sum_{\ell=1}^{N} [y_{\ell} \log(m_{\beta}(x^{\ell})) + (1-y_{\ell})\log(1 - m_{\beta}(x^{\ell}))] \label{ret2222}\\
\textrm{s.t.}\hspace{0.1cm} & \frac{1}{N}\displaystyle\sum_{\ell=1}^{N} \displaystyle (s_{\ell} - \bar{s})(\beta^\top x^{\ell}) \leq c\label{ret22221}\\
              & \frac{1}{N}\displaystyle\sum_{\ell=1}^{N} (s_{\ell} - \bar{s})(\beta^\top x^{\ell}) \geq - c\label{ret22222},
\end{align}
with
\begin{equation}\label{SensFea}
    \Bar{s} = \dfrac{\displaystyle\sum_{\ell=1}^N s_\ell}{N}
\end{equation}
being
$s$ the sensitive feature and $c \in \mathbb{R}^+$ is a threshold that controls the importance of fairness. However, if $c$ is chosen to be very small, the problem focuses exclusively on fairness, resulting in low accuracy.

In objective function \eqref{ret2222}, we have the original logistic regression objective function. Constraint \eqref{ret22221} and \eqref{ret22222}  guarantees the fairness in the classification. The construction and justification of these constraints can be found in \textcite{pmlr-v54-zafar17a}.

Now, using the logit link function, we can model a more specifically case of the GLMM problem and make it fair.

\subsection{Fair Generalized linear mixed model}
The goal of this section is to build upon existing results an algorithm that effectively handles both fairness considerations and the presence of random effects. This is done adjusting the Problem \eqref{ret2222} - \eqref{ret22222} to ensure that the model's classifications are not biased considering different stratas. We will discuss two proposals to address this problem. First, following the same strategy as \textcite{pamplona2024fair}, we encounter the following constrained optimization problem:

\begin{align}
\displaystyle \min_{\beta,b}\hspace{0.1cm} & -\displaystyle\sum_{i=1}^n \displaystyle\sum_{j=1}^{T_i} [y_{ij} \log(m_{\beta,b}(x^{ij})) + (1-y_{ij})\log(1 - m_{\beta,b}(x^{ij}))] + \lambda \sum_{i=1}^n b_i^2\label{ret12345}\\
\textrm{s.t.}\hspace{0.1cm} & \frac{1}{N}\displaystyle\sum_{i=1}^n \displaystyle\sum_{j=1}^{T_i} (s_{ij} - \bar{s})(\beta^\top x^{ij} + b_i) \leq c\label{ret122222222}\\
              & \frac{1}{N}\displaystyle\sum_{i=1}^n \displaystyle\sum_{j=1}^{T_i} (s_{ij} - \bar{s})(\beta^\top x^{ij} + b_i) \geq -c \label{ret1333233}
\end{align}
with
\begin{equation*}
    m_{\beta,b}(x^{ij}) = \dfrac{1}{1 + e^{-(\beta^\top x^{ij} + b_i)}},
\end{equation*}
$\Bar{s}$ from Equation \eqref{SensFea}. We denote the unconstrained Problem \eqref{ret12345} as Cluster-Regularized Logistic Regression (CRLR) and problem \eqref{ret12345} - \eqref{ret1333233} as Fair Cluster-Regularized Logistic Regression (Fair CRLR).

The second proposal is an adaptation of the boosting method proposed by \textcite{tutz2010generalized}. Boosting methods represent a powerful ensemble technique in machine learning, designed to sequentially combine weak learners into a strong predictive model. By iteratively fitting new models from the previous iteration. This process results in a highly accurate and robust model capable of handling complex patterns within the data \parencite{schapire1999brief}. Common boosting algorithms include AdaBoost \parencite{adaboost}, Gradient Boosting \parencite{gradboost}, and XGBoost \parencite{XGboost}.

Since the method proposed by \textcite{tutz2010generalized} is based on Newton's method \parencite{Lange2002}, it is crucial that the optimization problem is unconstrained. With this in mind, we convert this problem into an unconstrained problem using Lagrange's penalty, this strategy can be seen in \textcite{Nocedal2006}. To do this, we fix $c=0$, which means, Problem \eqref{ret12345} - \eqref{ret1333233} has a unique constraint. Observe that this is not a problem since we can control the constraint with the Lagrange multiplier $\rho$ allowing a penalized violation of it. So, we have the problem:
\begin{equation}\label{eq111}
\displaystyle \min_{\beta,b}\hspace{0.1cm} -\displaystyle\sum_{i=1}^n \displaystyle\sum_{j=1}^{T_i} [y_{ij} \log(m_{\beta,b}(x^{ij})) + (1-y_{ij})\log(1 - m_{\beta,b}(x^{ij}))] + \lambda \sum_{i=1}^n b_i^2 + \frac{\rho}{N} \Vert a^\top \delta \Vert^2_2
\end{equation}

with
$\delta^\top  = (\beta_0, \beta, b^\top) \in \mathbb{R}^{1+p+n}$
and 
\[
a^\top = \Big[\displaystyle\sum_{i=1}^n \displaystyle\sum_{j=1}^{T_i} (s_{ij} - \bar{s})[1 \quad (x^{ij})^\top] \quad  \displaystyle\sum_{j=1}^{T_i} (s_{1_j} - \bar{s}) \dots \displaystyle\sum_{j=1}^{T_i} (s_{n_j} - \bar{s}) \Big] \in \mathbb{R}^{1 \times (1+p+n)}.
\]

If we use this same strategy in Fair Logistic Regression, we can see that, by transforming the constraint into a penalization, it still respects the improvement in disparate impact, with results that are very similar to the original problem. This gives us another indication that the strategy works. The boxplots below were created using the same strategy that will be seen in Chapter \ref{sec:chapter-3}.

\begin{multicols}{2}
\begin{figure}[H]
\centering
\includegraphics[width=6.15cm]{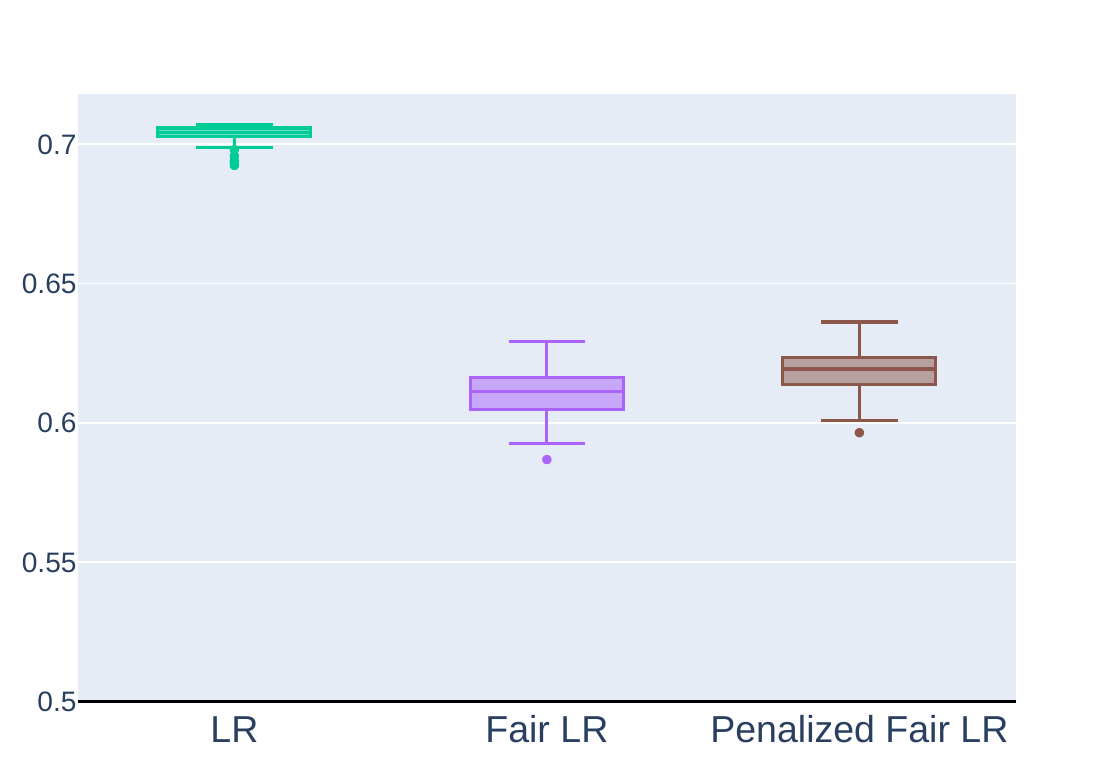}
\caption{Accuracy.}
\label{AC111111111}
\end{figure}

\begin{figure}[H]
\centering
\includegraphics[width=6.15cm]{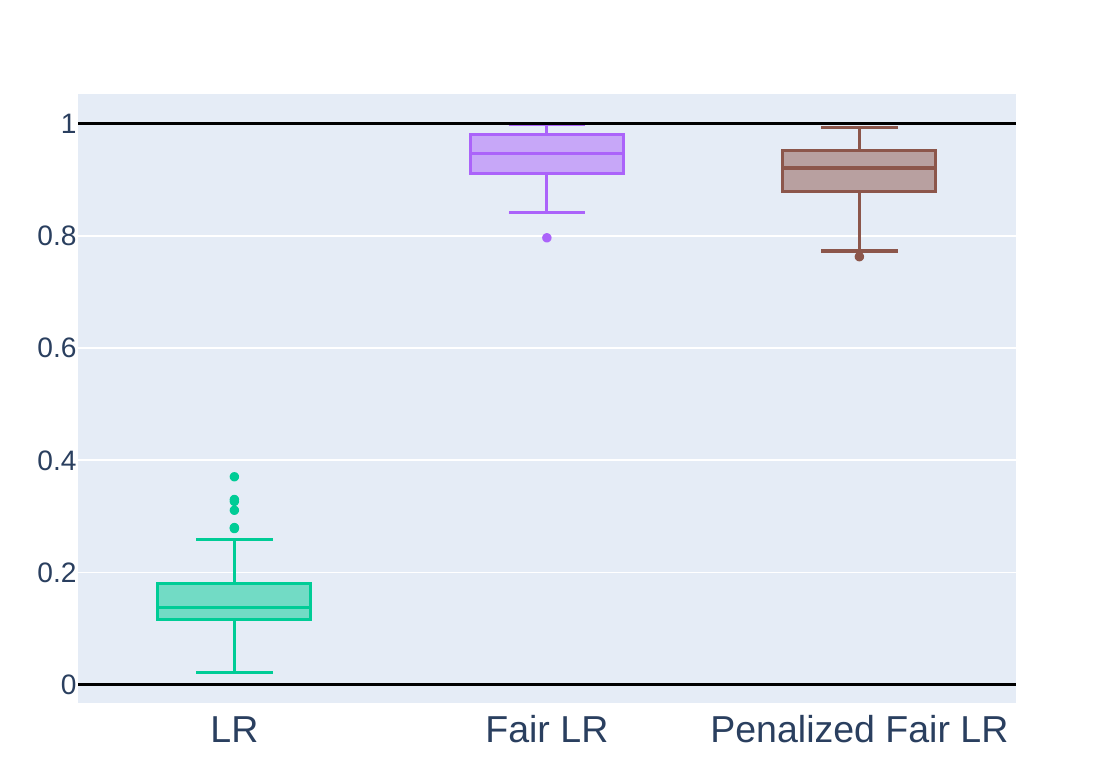}
\caption{Disparate impact.}
\label{DI111111111111}
\end{figure}
\end{multicols}

Similar to the propose of \textcite{tutz2010generalized} for solving the Problem \eqref{ret12345}, we do not explicitly solve the optimization problem \eqref{ret12345} - \eqref{ret1333233}. Instead, we propose an algorithm that, at each iteration, also updates the parameter $\lambda$, this stands in contrast to the traditional optimization problem where $\lambda$ is treated as a fix parameter.

We can now update the components that we will utilize in the iterative process of Fair Generalized Linear Mixed Models (Fair GLMMs), a type of GLMM designed to ensure fairness. In the following, we show in detail how to compute the components necessary for our algorithm based on the approach proposed by \textcite{tutz2010generalized}.

For $r \in [1, p]$ and for $l \in [1, l_{max}]$ with $l_{max}$ being the maximum number of iterations and the closed form for the pseudo Fisher matrix ($FP$)
\begin{equation}\label{FP}
    FP_r(\delta^{(l)}) = A_rW_lA_r + K  \in \mathbb{R}^{(n+2) \times (n+2)},
\end{equation}
with 
$A_r \define [X_r, Z] \in \mathbb{R}^{N \times (n+2)} $ ,
$K=Blockdiag(0,0, Q^{-1}, \dots, Q^{-1}) ~\in~ \mathbb{R}^{(n+2) \times (n+2)} $ being a block diagonal penalty matrix with diagonal of two zeros corresponding to intercept and the $r$-th fixed effect and $n$ times the matrix $Q^{-1}$, and given that we have a new objective function, we need to update the pseudo Fisher matrix.  By the Equation \eqref{eq111}, we have the penalization part being
\begin{equation}\label{penal}
    \dfrac{\rho}{N} \Vert a^\top \delta \Vert^2_2.
\end{equation}

As we can see in the maximization problem in \textcite{green1993nonparametric}, we can substitute the pseudo Fisher matrix with the negative of the Hessian matrix ($\mathcal{H}$). However, since we are working with a minimization problem, equality is automatically satisfied, i.e. $FP(\delta^{(l)}) = \mathcal{H}(\delta^{(l)})$. Hence, the Hessian of the penalization is:
\begin{equation}\label{eq15}
    \mathcal{H}(\dfrac{\rho}{N} \Vert a^\top \delta \Vert^2_2) = \dfrac{\rho}{N} a^\top a.
\end{equation}
Then we can obtain the final Hessian $(\mathcal{FH})$ of the objective function \eqref{eq111}  joining the Equations \eqref{FP} and \eqref{eq15} and considering,  w.l.o.g., $\rho = \dfrac{\rho}{N}$
\begin{equation}\label{FHess}\tag{FH}
\mathcal{FH}_r^{(l)} = A_rW_lA_r + K + \rho a_{ r}^\top a_{ r},
\end{equation}
with
\[
W_l = w(\delta^{(l-1)}),
\]
and
\[
w(\delta) = D(\delta)\Sigma^{-1}(\delta)D(\delta)^\top \in \mathbb{R}^{N \times N},
\]
where D is the derivative of the inverse of the link function. In our case we use the logit function, then $h(\eta^i) = g^{-1}(\eta^i) = e^{\eta^i}/(1 + e^{\eta^i})$, for $i \in [1, n]$ that is
\begin{align*}
D_i(\delta) &=  \frac{\partial h(\eta^i)}{\partial \eta} \\
 &=  \frac{e^{\eta^i}}{(1+e^{\eta^i})^2} \\
 &= \frac{e^{\eta^i} + e^{2\eta^i} - e^{2\eta^i}}{(1+e^{\eta^i})^2} \\
 &= \frac{e^{\eta^i}(1+e^{\eta^i}) - e^{2\eta^i}}{(1+e^{\eta^i})^2} \\
 &= \frac{(1+e^{\eta^i})(e^{\eta^i}(1+e^{\eta^i})-e^{2\eta^i})}{(1+e^{\eta^i})^3} \\
 &= \frac{e^{\eta^i}(1+e^{\eta^i})^2 - (1+e^{\eta^i})e^{2\eta^i}}{(1+e^{\eta^i})^3} \\
 &= \frac{e^{\eta^i}}{1+e^{\eta^i}} - \frac{e^{2\eta^i}}{(1+e^{\eta^i})^2} \\
 &=  \Big(\frac{e^{\eta^i}}{1 + e^{\eta^i}}\Big) \Big( 1 - \frac{e^{\eta^i}}{1 + e^{\eta^i}}\Big)
 \end{align*}
\begin{align}\label{D}
\hspace{-0.97cm} = h(\eta^i)(1-h(\eta^i)).
\end{align}
Moreover, by \textcite{Breslow1993},
\begin{align}\label{sigma}
    \Sigma_i(\delta) = cov(y_i|\beta,b_i)  = h(\eta^i)(1-h(\eta^i))  \in \mathbb{R}^{T_i \times T_i} .
\end{align}
Combining \eqref{D} and \eqref{sigma} we obtain
\[
D(\delta) = \Sigma(\delta),
\]
being $D = diag(D_1, \dots, D_n)$, and then
\begin{align*}
    w(\delta) &= D(\delta)\Sigma^{-1}(\delta)D(\delta)^\top  
    = D(\delta)D^{-1}(\delta)D(\delta)^\top  \\
    &= D(\delta)^\top  \in \mathbb{R}^{N \times N} .
\end{align*}

For the score function for $r \in [1, p]$, we have the closed form obtained by differentiating the objective function \eqref{ret12345}
\begin{equation}\label{s}
    s_r(\delta^{(l)}) = A_r^\top W_lD_l^{-1}(y-\mu^{(l)})- K\delta_r^{(l)}  \in \mathbb{R}^{(n+2) \times 1}
\end{equation}
being $\mu^{(l)} = h(\eta^{(l)}) \in \mathbb{R}^{N} $, $\delta_r^\top = (\beta_0, \beta_r, b^\top)$ and $D_l = D(\delta^{(l-1)})$ that come by differentiating the inverse of the logit function. Furthermore, the score function \eqref{ret12345} can be interpreted as the negative of the gradient $(\nabla)$, for the same reason as the Hessian, as it is calculated based on a maximization problem which we adapt here to a minimization problem. Therefore, the negative of the gradient of the penalization part \eqref{penal} is:
\begin{equation}\label{eq14}
-\nabla(\dfrac{\rho}{N} \Vert a^\top \delta \Vert^2_2) = -2 \dfrac{\rho}{N} (a^\top a) \delta,
\end{equation}
to obtain the final score function $(\mathcal{FS})$ of the objective function \eqref{eq111}  we join the Equations \eqref{s} and \eqref{eq14} and considering,  w.l.o.g., $\rho = 2\dfrac{\rho}{N}$
\begin{equation}\label{FScore}\tag{FS}
\mathcal{FS}_r^{(l)} = s_r(\delta^{(l-1)}) - \rho a_{ r}^\top(a_{ r} \delta^{(l-1)}_r).
\end{equation}

For finding the best direction for the update we use the Bayesian Information Criterion $(BIC) \in \mathbb{R}^{p}$. The $BIC$ is a popular model selection criterion for GLMMs for being relatively easy to calculate, and it has been shown to perform well in a variety of simulations as can be seen in \textcite{vrieze2012model}:
\begin{equation*}
    BIC_r^{(l)} = -2\Omega(\mu_r^{(l)}) + 2tr(H_r^{(l)})log(n),
\end{equation*}
with
\[
\Omega(\mu_r^{(l)}) = \displaystyle\sum_{i=1}^n \displaystyle\sum_{j=1}^{T_i} y_{ij}log(\mu_{ij}) + (1-y_{ij})(1-log(\mu_{ij})) - \rho \Vert a^\top \delta_r^{(l)} \Vert^2_2,
\]
that comes from the objective function, without considering the variance penalization of the random effects, as this is already considered in other stages of the process, and the hat matrix corresponding to the $l$-th boosting step considering the $r$-th component,
\[
H_r^{(l)} = I - (I - M_r^{(l)})(I - M_{l-1})(I - M_{l-2}) \dots (I - M_0) \in \mathbb{R}^{N \times N},
\]
with $M_k$ the matrix corresponding to the component that has been selected in the $k$-th iteration, for $k = 1, \dots, l-1$ being
\[
M_r^{(l)} \define \Sigma_l^{1/2}\tilde{H}_r^{(l)} \Sigma_l^{-1/2} \in \mathbb{R}^{N \times N},
\]
and
\[
M_0 \define A_1(A_1^\top W_1 A_1 + K_1)A_1^\top W_1 \in \mathbb{R}^{N \times N},
\]
being $\Sigma_l = \Sigma(\delta^{(l-1)})$, and an adaptation of the generalized ridge regression hat-matrix which considers the penalization of fairness constraints,
\[
\tilde{H}_r^{(l)} = W_l^{1/2} A_r (A_rW_lA_r + K + \rho \Vert a^\top \delta_r^{(l)} \Vert^2_2)^{-1} A_r^\top W_l^{1/2} \in \mathbb{R}^{N \times N}.
\]

Now, as can be seen in \textcite{harville1977maximum} we can update the covariance matrix $Q^{(l)}$ by
\begin{equation*}
    Q^{(l)} = \frac{1}{n} \displaystyle\sum_{i=1}^n (V_{i}^{(l)} + b_i^{(l)}(b_i^{(l)})^\top ) \in \mathbb{R}.
\end{equation*}
In general, and in our case, the $V_{i}$ are computed via the formula
\[
V_{i} = F_{i}^{-1} + F_{i}^{-1} \tilde{F}_i^\top(\hat{F} - \displaystyle\sum_{i=1}^n \tilde{F}_i F_{i}^{-1} \tilde{F}_i^\top)^{-1} \tilde{F}_iF_{i}^{-1} \in \mathbb{R}
\]
with
\[
F_{i} = (Z^{i})^\top  D_i(\delta) \Sigma_i(\delta)^{-1} D_i(\delta) Z^{i} + Q^{-1} \in \mathbb{R},
\]
\[
\tilde{F}_i = (X^{i})^\top  D_i(\delta) \Sigma_i(\delta)^{-1} D_i(\delta) Z^{i} \in \mathbb{R}^{p},
\]
and
\[
\hat{F} = \displaystyle\sum_{i=1}^n (X^{i})^\top  D_i(\delta) \Sigma_i(\delta)^{-1} D_i(\delta) X^{i} \in \mathbb{R}^{p \times p},
\]
where $\hat{F}$, $\tilde{F}_i$ and $F_{i}$ are the elements of the pseudo Fisher matrix $FP(\delta)$ of the full model. For more detailed derivation see \textcite{gu2012generalized}.

Preliminary tests have shown that model \eqref{ret12345} - \eqref{ret1333233} is an warm starting for algorithm $1$. Consequently, we have the following initial parameters:
\begin{itemize}
    \item $\beta_0^{(0)}$ from Fair CRLR;
    \item $\beta^{(0)}$ from Fair CRLR;
    \item $b^{(0)}$  from Fair CRLR;
    \item $\mu^{(0)} = 0 \in {\mathbb{R}}^N$;
    \item $\eta^{(0)} = 0 \in {\mathbb{R}}^N$; 
    \item $Q^{(0)} = 2.0$.
\end{itemize}

Finally, we have all necessary components and motivations to propose an algorithm for solving the Fair Generalized Linear Mixed Model. 
\newpage
\begin{table}[ht]\label{ALGO1}
	\centering
	\begin{tabular}{|l|}\hline
		\;\;\;\;\;\textbf{Algorithm 1} : Fair Generalized Linear Mixed Model
		\;
		\\ 	\hline
		\textbf{Given}:
            $\mu^{(0)}$,
            $\beta_0^{(0)}$,
             $\beta^{(0)}$,
            $b^{(0)}$,
            $\eta^{(0)}$,
            $l_{max}$,
            $Q^{(0)}$. \\
  
		\textbf{Iteration}: \\
		\hspace*{0.15cm} \begin{tabular}{|l}
			\textbf{(1)  Refitting of residuals}: \\
			For $l \in [1, l_{max}]$, \\
			\hspace*{0.15cm} \begin{tabular}{|l}
			 \textbf{(i) Computation of parameters} \\
                 For $r \in [1, p]$ the $r$-th Newton's method step has the form: \\
            
                 \hspace*{3cm}$\delta^{(l)}_r = (\mathcal{FH}_r^{(l-1)})^{-1}(\mathcal{FS}_r^{(l-1)})$
            \\

                 \textbf{(ii) Selection step} \\
                Select from $r \in [1, p]$ the index $j$ corresponding to the smallest $BIC_r^{(l)}$ \\ and select the related $(\delta^{(l)}_j)^\top  = (\beta_0^*, \beta_j^*, (b^*)^\top )$.\\

                 \textbf{(iii) Update} \\
                Set \\
                    $
                    \hspace*{3cm}\beta^{(l)}_0 = \beta^{(l-1)}_0 + \beta^*_0, 
                    $ \\
                    and \\
                    $
                    \hspace*{3cm}b^{(l)} = b^{(l-1)} + b^{*}
                    $ \\
                    and for $r \in [1, p]$ set \\
                    \hspace*{3cm}$ \beta_r^{(l)} =
                        \begin{cases}
                          \beta_r^{(l-1)} \hspace{1.5cm} \text{if } r \neq j\\
                          \beta_r^{(l-1)} + \beta_r^* \hspace{0.7cm} \text{if } r = j
                        \end{cases}\,
                    $ \\
                with $A \define [X,Z]$ update \\
                $
                \hspace*{3cm}\eta = A\delta^{(l)}
                $
			\end{tabular} \\
            \textbf{(2) Computing of variance-covariance components}: \\
                \hspace*{3.47cm}$
                    Q^{(l)} = \dfrac{1}{n} \displaystyle\sum_{i=1}^n (V_{i}^{(l)} + b_i^{(l)}(b_i^{(l)})^\top ).
                $
            
		\end{tabular} \\
		until $Q^{(l)} = Q^{(l-1)}$.
		\\ \hline
	\end{tabular}
\end{table}

The following section presents various numerical tests demonstrating the efficacy of our proposed methods.

%% file: chapter4.tex
\section{Simulation Study}\label{sec:chapter-3}

In this section, we aim to present numerical tests to demonstrate the effectiveness of the proposed method.

First we present the step-by-step strategy used to create the datasets and to conduct the numerical experiments. Using the \texttt{Julia 1.9} \parencite{BezansonEdelmanKarpinskiShah17} language with the packages \texttt{Distributions} \parencite{distributions}, \texttt{GLM} \parencite{GLM},  \texttt{MixedModels} \parencite{MixedModels}, \texttt{DataFrames} \parencite{DataFrames}, \texttt{MLJ} \parencite{MLJ} and \texttt{MKL} \parencite{MKL}, we generate the following parameters:

\begin{itemize}
        \item \emph{Number of points}: Number of points in the dataset; 
        \item \emph{$\beta 's$}: The fixed effects;
        \item \emph{$b's$}: The random effects with distribution $N(0,Q)$, with covariance matrix $Q$ (Used only when indicated);
        \item \emph{Data points}: The covariate vector associated with fixed effects with distribution $N(0,1)$;
        \item $c$: Threshold from Fair Logistic Regression;
        \item $\rho$: Lagrange multiplier;
        \item \emph{seed}: Random seed used in the generation of data;
        \item \emph{Train-Test split}: Approximately 0.4\% of the dataset was used for the training set, and 99.6\% for the test set. This percentage was due to the fact that we randomly selected 3 to 5 points from each strata for the training set.
\end{itemize}
The labels of the synthetic dataset were computed using
\begin{equation}\label{m1}
    m = \dfrac{1}{1 + e^{-(\beta^T x + b)}}
\end{equation}
in tests where the dataset has random effects, and
\begin{equation}\label{m2}
    m = \dfrac{1}{1 + e^{-\beta^T x}}
\end{equation}
in tests where the dataset has just fixed effects. Finally,
\begin{equation*}
    y = Binomial(1, m).
\end{equation*}

The comparisons are be made by comparing the accuracy and the disparate impact between the Algorithms:\begin{enumerate}
    \item Generalized linear mixed model (GLMM);
    \item Fair generalized linear mixed model (Fair GLMM);
    \item Cluster-Regularized Logistic Regression (CRLR);
    \item Fair Cluster-Regularized Logistic Regression (Fair CRLR),
    \item Logistic Regression (LR);
    \item Fair Logistic Regression (Fair LR).
\end{enumerate}
The test were conducted on a laptop with an Intel Core i9-13900HX processor with a clock speed of 5.40 GHz, 64 GB of RAM, and Windows 11 operating system, with 64-bit architecture.

To compute accuracy, first we need to compute the classifications using Equations \eqref{m1} and \eqref{m2} with,
\begin{equation*} \hat{y} =
                \begin{cases}
                  1 \hspace{0.4cm} \text{if } \hspace{0.05cm} m \geq 0.5\\
                  0 \hspace{0.4cm} \text{if } \hspace{0.05cm} m < 0.5
                \end{cases}\,.
\end{equation*}
Given the true label of all points, we can distinguish them into four categories: true positive (TP) or true negative (TN) if the point is classified correctly in the positive or negative class, respectively, and false positive (FP) or false negative (FN) if the point is misclassified in the positive or negative class, respectively. Based on this, we can compute the accuracy, where a higher value indicates a better classification, as follows,
\begin{equation*}
    AC = \dfrac{TP + TN}{TP + TN + FP + FN} \in [0,1].
\end{equation*}

To compute the disparate impact of a specific sensitive feature $s$ we use the following equation as can be seen in \textcite{DIII},
\begin{equation*}
    DI = \dfrac{p(\hat{y}|s=1)}{p(\hat{y}|s=0)} \in [0,\infty).
\end{equation*}

Disparate impact, as a measure, should be equal to 1. This indicates that discrimination does not exist. Values greater or lower than 1 suggest that unwanted discrimination exists. However, $DI = 2$ and $DI = 0.5$ represent the same level of discrimination, although in the first case, the difference between the perfect value is 1, and in the latter case, it is 0.5. To avoid such situations, we use the minimum of $DI$ and its inverse.

\begin{equation*}
    DI = \min \Big( \dfrac{p(\hat{y}|s=1)}{p(\hat{y}|s=0)}, \dfrac{p(\hat{y}|s=0)}{p(\hat{y}|s=1)} \Big)  \in [0,1].
\end{equation*}

In the following, we generate four different synthetic populations (scenarios) to compare the competing algorithms. For each synthetic data set 100 samples are drawn. The simulation results are discussed for each synthetic data set using 2 images that represent, respectively, the accuracy and the disparate impact.

All figures were created using the \texttt{Plots}  and \texttt{PlotlyJS} packages, developed by \textcite{Plots} and all hyperparameters were selected via cross-validation \parencite{browne2000cross}. All tests can be reproduced, and the codes of all functions used can be found in \href{https://github.com/JoaoVitorPamplona/Fair-Generalized-Linear-Mixed-Models}{GitHub}.

\subsection{Unfair population with strata effect}\label{scenario3.1}
Parameters of data generation:
\begin{multicols}{2}
\begin{itemize}
    \item $\beta$'s $= [-2.0, 0.4, 0.8, 0.5, 3.0]$;
    \item $b$'s: 100 stratas with $b_i~\sim~N(0,3.0)$, with $i \in [1, 100]$;
    \item $c = 0.1$;
    \item $\rho = 0.8$;
    \item $\lambda = 1$.
\end{itemize}
\end{multicols}
\begin{table}[ht]
\centering
\begin{tabular}{|c c c c c c c|}
\hline
Algorithm                & Mean & p25 & Median& p75  & p95  & std  \\ \hline
GLMM                    & 0.89 & 0.88 & 0.89 & 0.90 & 0.91 & 0.01              \\ \hline

Fair GLMM               & 0.82 & 0.82 & 0.82 & 0.83 & 0.83 & 0.01                 \\ \hline

CRLR                    & 0.79 & 0.78 & 0.79 & 0.80 & 0.81 & 0.02                \\ \hline

Fair CRLR               & 0.82 & 0.81 & 0.82 & 0.83 & 0.86 & 0.02                 \\ \hline

LR                       & 0.68 & 0.68 & 0.68 & 0.68 & 0.68 & 0.01                 \\ \hline

Fair LR                 & 0.64 & 0.63 & 0.64 & 0.64 & 0.66 & 0.01                 \\ \hline
\end{tabular}
\caption{Accuracy.}
\label{tabela5}
\end{table}

\begin{table}[ht]
\centering
\begin{tabular}{|c c c c c c c|}
\hline
Algorithm                & Mean & p25 & Median& p75  & p95  & std  \\ \hline
GLMM                    & 0.48 & 0.45 & 0.48 & 0.51 & 0.56 & 0.05              \\ \hline

Fair GLMM               & 0.91 & 0.90 & 0.91 & 0.92 & 0.94 & 0.02                 \\ \hline
CRLR                     & 0.18 & 0.14 & 0.18 & 0.22 & 0.27 & 0.06                \\ \hline

Fair CRLR                & 0.66 & 0.63 & 0.66 & 0.70 & 0.75 & 0.05                 \\ \hline
LR                       & 0.10 & 0.04 & 0.09 & 0.14 & 0.21 & 0.07                 \\ \hline

Fair LR                  & 0.52 & 0.48 & 0.53 & 0.59 & 0.65 & 0.11                 \\ \hline
\end{tabular}
\caption{Disparate impact}
\label{tabela6}
\end{table}

\newpage
\begin{multicols}{2}
\begin{figure}[H]
\centering
\includegraphics[width=6.15cm]{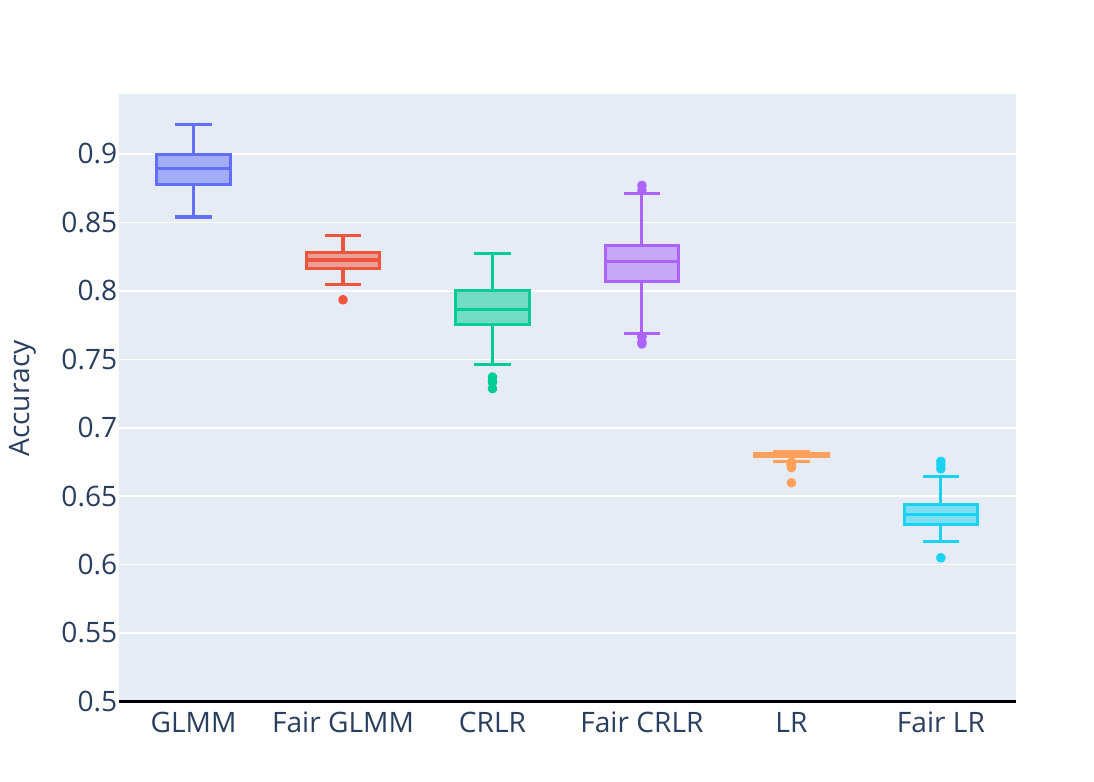}
\caption{Accuracy.}
\label{AC3}
\end{figure}

\begin{figure}[H]
\centering
\includegraphics[width=6.15cm]{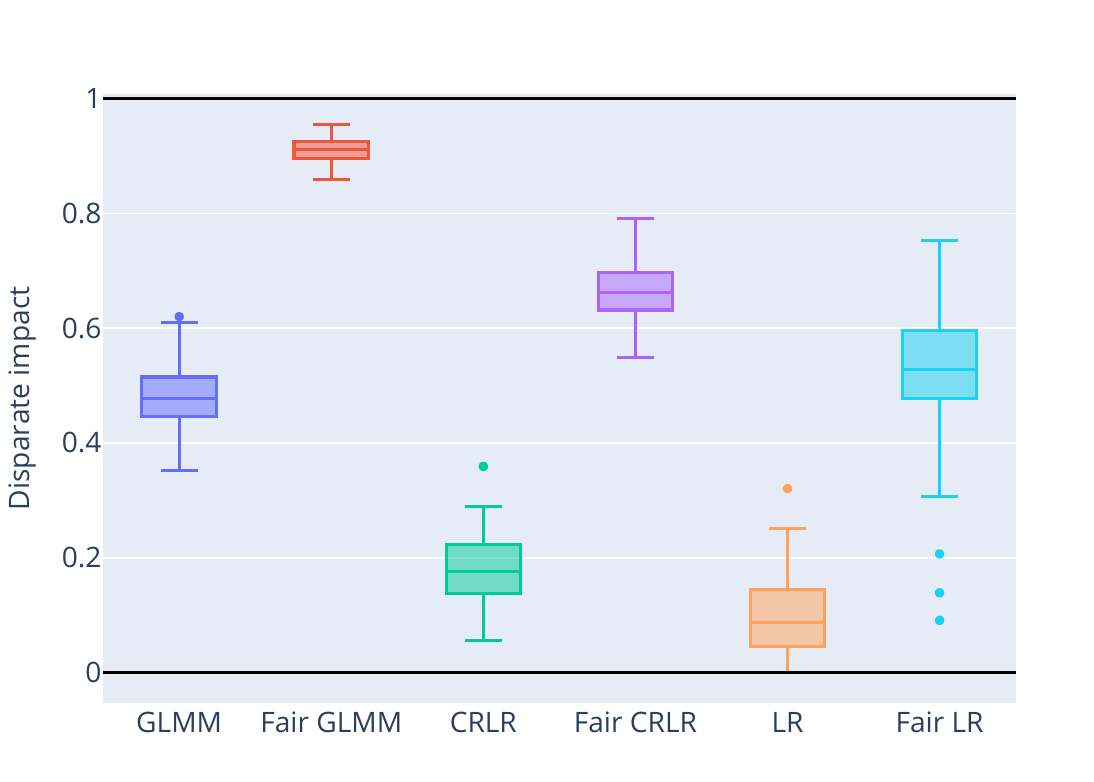}
\caption{Disparate impact.}
\label{DI3}
\end{figure}
\end{multicols}

The results in Table \ref{tabela5} and Figure \ref{AC3} demonstrate superior accuracy for GLMM, Fair GLMM, CRLR, and Fair CRLR in this experiment set. This is unsurprising, as logistic regression doesn't account for random effects. We can also see that the GLMM performs better on both metrics compared to the optimization problems.

Table \ref{tabela6} and Figure \ref{DI3} show that we also obtained an improvement in the disparate impact on the Fair algorithms. Note that make sense since we have a unfair population.


\subsection{Fair population with strata effect}
Parameters of data generation:
\begin{multicols}{2}
\begin{itemize}
     \item $\beta$'s $= [-0.1, 1, 1, 1, 0.1]$;
    \item $b$'s: 100 stratas with $b_i~\sim~N(0,3.0)$, with $i \in [1, 100]$;
    \item $c = 0.1$;
    \item $\rho = 0.8$;
    \item $\lambda = 1$.
\end{itemize}
\end{multicols}
\begin{table}[ht]
\centering
\begin{tabular}{|c c c c c c c|}
\hline
Algorithm                & Mean & p25 & Median& p75  & p95  & std  \\ \hline
GLMM                    & 0.87 & 0.86 & 0.89 & 0.90 & 0.90 & 0.06              \\ \hline

Fair GLMM               & 0.87 & 0.87 & 0.88 & 0.88 & 0.89 & 0.02                 \\ \hline
CRLR                     & 0.81 & 0.80 & 0.81 & 0.82 & 0.83 & 0.02                \\ \hline

Fair CRLR                & 0.81 & 0.80 & 0.81 & 0.82 & 0.83 & 0.02                 \\ \hline
LR                       & 0.66 & 0.66 & 0.66 & 0.66 & 0.66 & 0.01                 \\ \hline

Fair LR                 & 0.66 & 0.66 & 0.66 & 0.66 & 0.66 & 0.01                 \\ \hline
\end{tabular}
\caption{Accuracy.}
\label{tabela3}
\end{table}

\begin{table}[ht]
\centering
\begin{tabular}{|c c c c c c c|}
\hline
Algorithm                & Mean & p25 & Median& p75  & p95  & std  \\ \hline
GLMM                    & 0.96 & 0.94 & 0.97 & 0.99 & 0.99 & 0.03              \\ \hline

Fair GLMM               & 0.98 & 0.97 & 0.98 & 0.99 & 0.99 & 0.02                 \\ \hline
CRLR                     & 0.91 & 0.86 & 0.92 & 0.96 & 0.98 & 0.07                \\ \hline

Fair CRLR                & 0.92 & 0.88 & 0.93 & 0.96 & 0.98 & 0.05                 \\ \hline
LR                       & 0.88 & 0.82 & 0.90 & 0.95 & 0.99 & 0.09                 \\ \hline

Fair LR                 & 0.89 & 0.84 & 0.91 & 0.95 & 0.99 & 0.07                 \\ \hline
\end{tabular}
\caption{Disparate impact}
\label{tabela4}
\end{table}

\newpage
\begin{multicols}{2}
\begin{figure}[H]
\centering
\includegraphics[width=6.15cm]{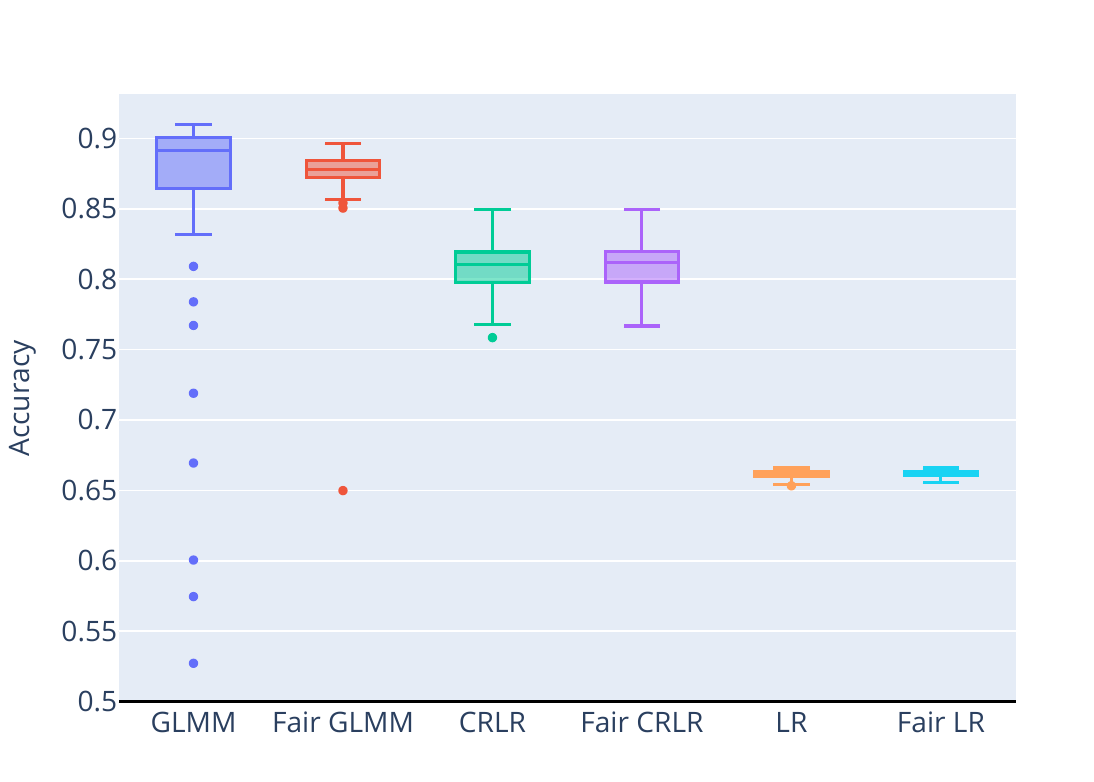}
\caption{Accuracy.}
\label{AC2}
\end{figure}

\begin{figure}[H]
\centering
\includegraphics[width=6.15cm]{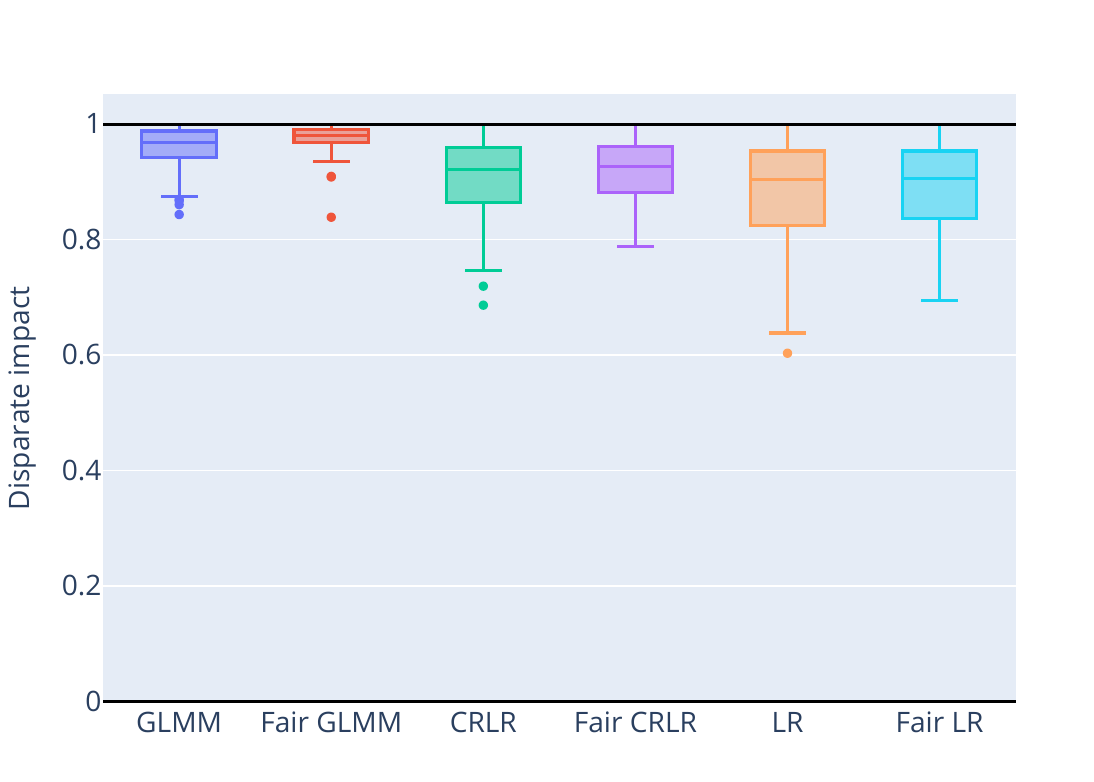}
\caption{Disparate impact.}
\label{DI2}
\end{figure}
\end{multicols}

As can be seen in Table \ref{tabela3} and in the Figure \ref{AC2}, in this set of experiments, we obtained a better accuracy in GLMM, Fair GLMM, CRLR and Fair CRLR. This is expected since logistic regression does not account for random effects.

Table \ref{tabela4} and Figure \ref{DI2} show that the disparate impact remains almost the same in all tests since we have a fair population.


\subsection{Unfair population without strata effect}\label{ssUw}
Parameters of data generation:
\begin{multicols}{2}
\begin{itemize}
    \item $\beta$'s $= [-2.0, 0.4, 0.8, 0.5, 3.0]$;
    \item $c = 0.1$;
    \item $\rho = 0.8$;
    \item $\lambda = 1$.
\end{itemize}
\end{multicols}
\begin{table}[ht]
\centering
\begin{tabular}{|c c c c c c c|}
\hline
Algorithm                & Mean & p25 & Median& p75  & p95  & std  \\ \hline
GLMM                     & 0.70 & 0.70 & 0.71 & 0.72 & 0.73 & 0.03              \\ \hline

Fair GLMM                & 0.61 & 0.61 & 0.62 & 0.62 & 0.63 & 0.01                 \\ \hline
CRLR                     & 0.79 & 0.78 & 0.79 & 0.79 & 0.79 & 0.01                \\ \hline

Fair CRLR                & 0.68 & 0.67 & 0.68 & 0.69 & 0.70 & 0.01                 \\ \hline
LR                       & 0.79 & 0.79 & 0.79 & 0.79 & 0.79 & 0.01                 \\ \hline

Fair LR                  & 0.69 & 0.68 & 0.69 & 0.70 & 0.71 & 0.01                 \\ \hline
\end{tabular}
\caption{Accuracy.}
\label{tabela9}
\end{table}

\begin{table}[ht]
\centering
\begin{tabular}{|c c c c c c c|}
\hline
Algorithm                & Mean & p25 & Median& p75  & p95  & std  \\ \hline
GLMM                    & 0.17 & 0.12 & 0.15 & 0.17 & 0.25 & 0.15              \\ \hline

Fair GLMM               & 0.73 & 0.70 & 0.74 & 0.78 & 0.82 & 0.07                 \\ \hline
CRLR                    & 0.04 & 0.02 & 0.04 & 0.05 & 0.08 & 0.02               \\ \hline

Fair CRLR               & 0.56 & 0.51 & 0.57 & 0.62 & 0.67 & 0.08                 \\ \hline
LR                      & 0.04 & 0.02 & 0.03 & 0.05 & 0.07 & 0.02                 \\ \hline

Fair LR                 & 0.52 & 0.47 & 0.53 & 0.59 & 0.64 & 0.09                 \\ \hline
\end{tabular}
\caption{Disparate impact}
\label{tabela10}
\end{table}

\newpage
\begin{multicols}{2}
\begin{figure}[H]
\centering
\includegraphics[width=6.15cm]{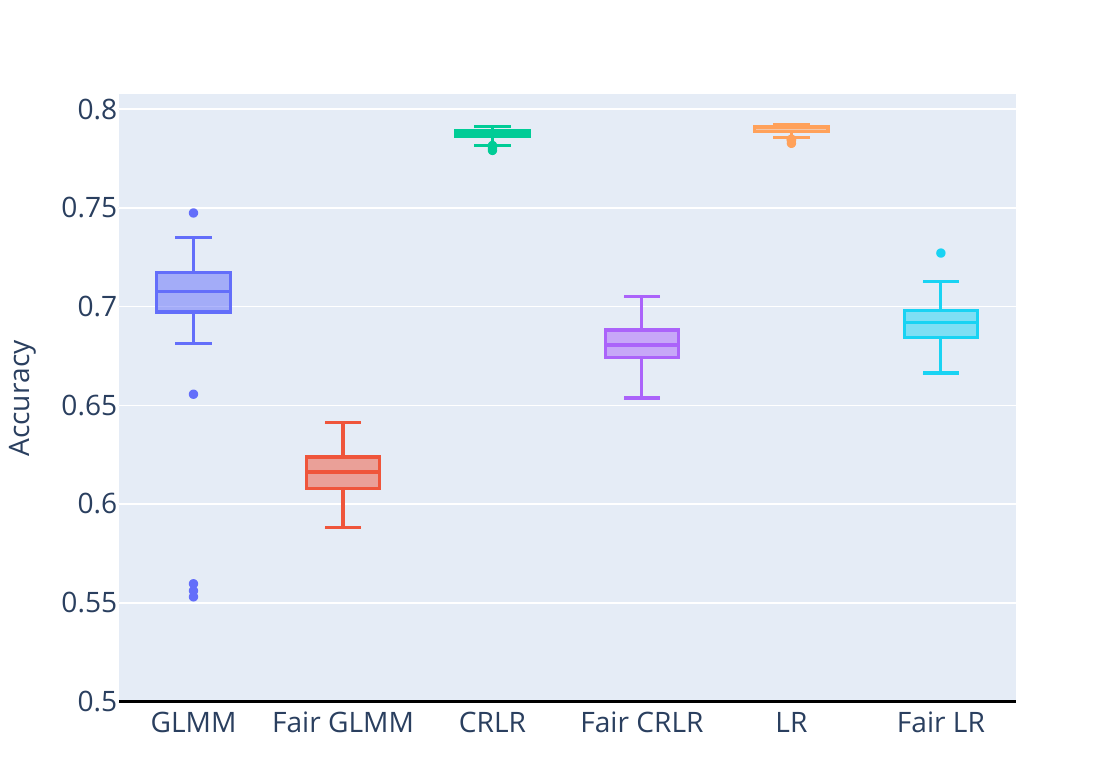}
\caption{Accuracy.}
\label{AC5}
\end{figure}

\begin{figure}[H]
\centering
\includegraphics[width=6.15cm]{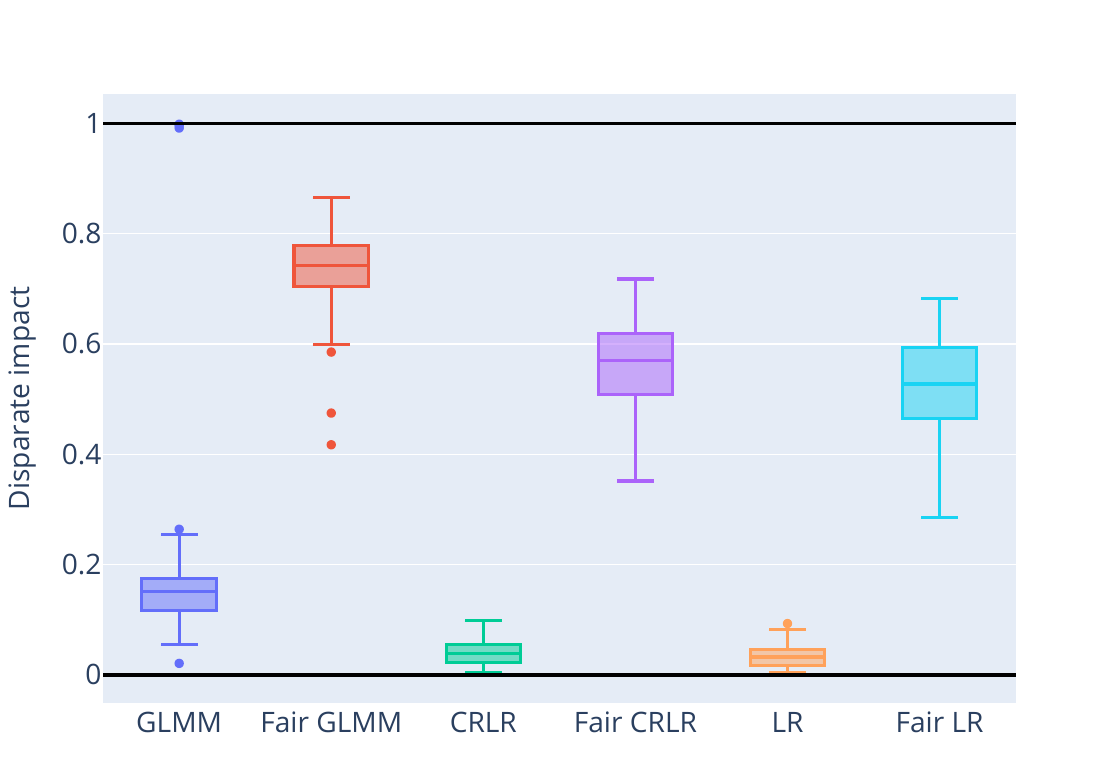}
\caption{Disparate impact.}
\label{DI5}
\end{figure}
\end{multicols}

Our experiments, detailed in Table \ref{tabela9} and Figure \ref{AC5}, show a slight decrease in accuracy for GLMM algorithms. This is likely because GLMMs attempt to account for random effects, even when they are absent. This additional parameter estimation in GLMMs can lead to a reduction in accuracy compared to logistic regression optimization problems.

Table \ref{tabela10} and Figure \ref{DI5} show that we obtained an improvement in the disparate impact on the Fair algorithms. Note that make sense since we have a unfair population.


\subsection{Fair population without strata effect}\label{ssUw2}
Parameters of data generation:
\begin{multicols}{2}
\begin{itemize}
    \item $\beta$'s $= [-0.1, 1, 1, 1, 0.1]$;
    \item $c = 0.1$;
    \item $\rho = 0.8$;
    \item $\lambda = 1$.
\end{itemize}
\end{multicols}

\begin{table}[ht]
\centering
\begin{tabular}{|c c c c c c c|}
\hline
Algorithm                & Mean & p25 & Median& p75  & p95  & std  \\ \hline
GLMM                     & 0.63 & 0.61 & 0.63 & 0.67 & 0.68 & 0.04              \\ \hline

Fair GLMM                & 0.67 & 0.67 & 0.68 & 0.68 & 0.69 & 0.02                 \\ \hline
CRLR                     & 0.75 & 0.75 & 0.75 & 0.75 & 0.75 & 0.01                \\ \hline

Fair CRLR                & 0.75 & 0.75 & 0.75 & 0.75 & 075 & 0.01                 \\ \hline
LR                       & 0.75 & 0.75 & 0.75 & 0.75 & 0.76 & 0.01                 \\ \hline

Fair LR                  & 0.75 & 0.75 & 0.75 & 0.75 & 0.76 & 0.01                 \\ \hline
\end{tabular}
\caption{Accuracy.}
\label{tabela7}
\end{table}

\begin{table}[ht]
\centering
\begin{tabular}{|c c c c c c c|}
\hline
Algorithm                & Mean & p25 & Median& p75  & p95  & std  \\ \hline
GLMM                     & 0.88 & 0.82 & 0.89 & 0.95 & 0.99 & 0.09              \\ \hline

Fair GLMM                & 0.94 & 0.91 & 0.94 & 0.97 & 0.99 & 0.04                 \\ \hline
CRLR                     & 0.90 & 0.85 & 0.90 & 0.95 & 0.99 & 0.07                \\ \hline

Fair CRLR                & 0.91 & 0.86 & 0.91 & 0.95 & 0.99 & 0.06                 \\ \hline
LR                       & 0.89 & 0.85 & 0.90 & 0.94 & 0.99 & 0.06                 \\ \hline

Fair LR                  & 0.91 & 0.86 & 0.91 & 0.95 & 0.99 & 0.06                 \\ \hline
\end{tabular}
\caption{Disparate impact}
\label{tabela8}
\end{table}

\begin{multicols}{2}
\begin{figure}[H]
\centering
\includegraphics[width=6.15cm]{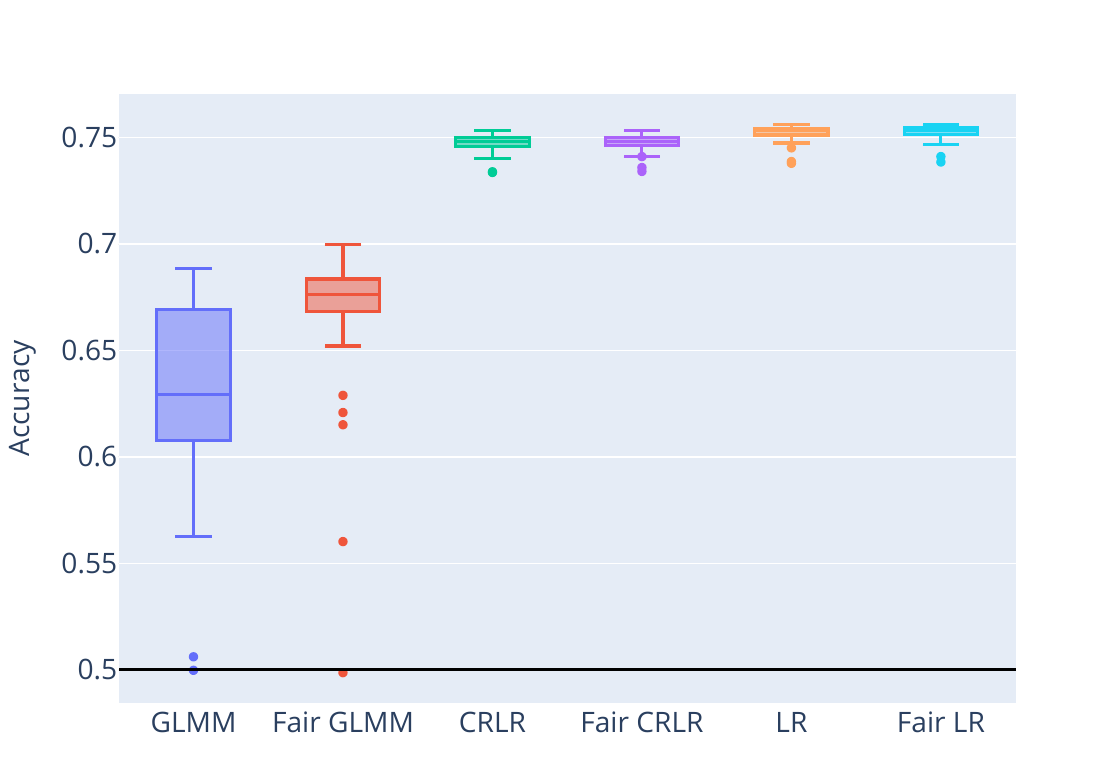}
\caption{Accuracy.}
\label{AC4}
\end{figure}

\begin{figure}[H]
\centering
\includegraphics[width=6.15cm]{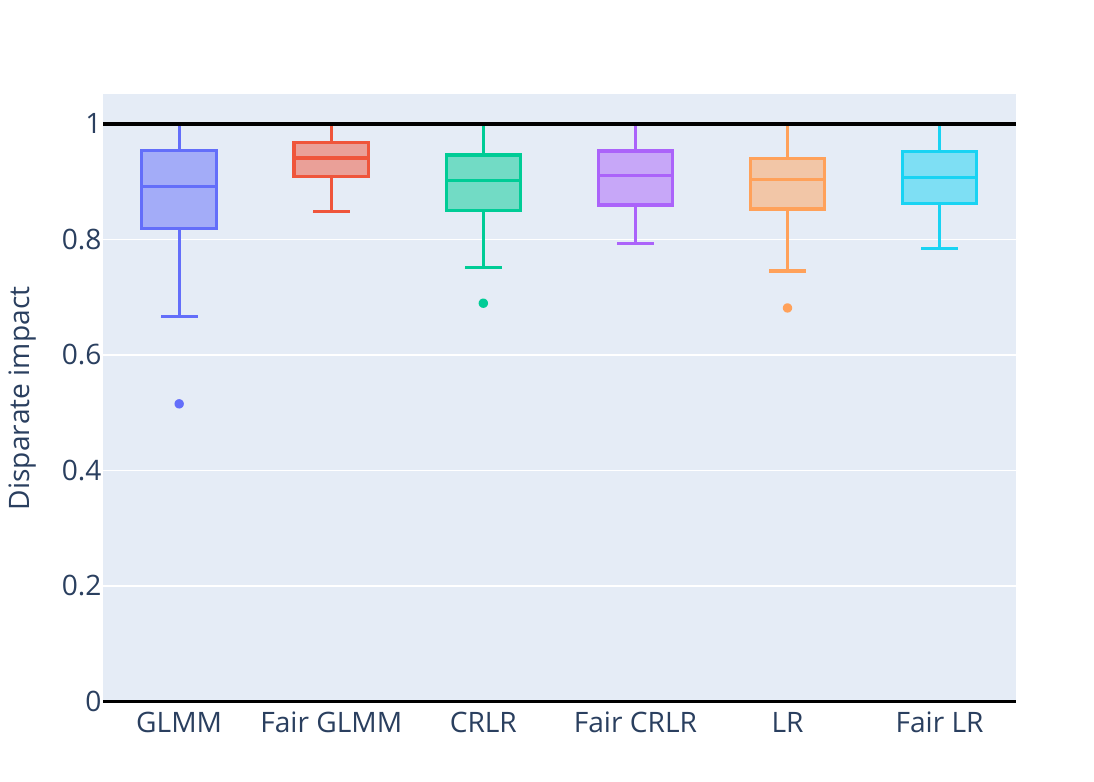}
\caption{Disparate impact.}
\label{DI4}
\end{figure}
\end{multicols}

Our experiments, detailed in Table \ref{tabela7} and Figure \ref{AC4}, show a decrease in accuracy for GLMM algorithms for the same reason from Scenario \ref{ssUw}.

Table \ref{tabela8} and Figure \ref{DI4} show that we also have a very similar disparate impact in all algorithms. Note that make sense since we have a fair population.


%% file: chapter5.tex
\section{Application}\label{sec:chapter-5}

In this set of experiments, we test the Bank marketing dataset, which has a strata bias related to the duration of telephone calls, the longer the call duration, (calls with longer duration imply a higher probability of the label being $1$), and a sensitive feature, which in this case is housing loan, the housing loan feature can be considered a sensitive feature because it is directly linked to injustice in its generation, as can be seen in \textcite{sitefair} and in \textcite{unfair}. The data is related to direct marketing campaigns of a Portuguese banking institution. The marketing campaigns were based on phone calls. Often, more than one contact was required with the same client, in order to determine if the product (bank term deposit) would be subscribed ($y = 1$) or not ($y = 0$).  Were conducted 100 samples with 3.5\% of the data as the training set and the remaining data as the test set. Since the application under study employs random effects, we conduct the comparisons using GLMM, the cluster-regularized logistic regression and the regular logistic regression.

The features used in the classification process are the follows:

\begin{multicols}{2}
\begin{itemize}
    \item Age
    \item Job
    \item Marital status
    \item Education
    \item Has credit in default?
    \item Has housing loan?
    \item Has personal loan?
    \item Contact communication type
    \item Last contact month of year
    \item Last contact day of the week
    \item Number of contacts performed during this campaign and for this client
    \item Number of days that passed by after the client was last contacted from a previous campaign
    \item Number of contacts performed before this campaign and for this client 
    \item Outcome of the previous marketing campaign
    \item Employment variation rate
    \item Consumer price index
    \item Consumer confidence index
    \item Euribor 3 month rate 
    \item Number of employees
\end{itemize}
\end{multicols}

\begin{table}[ht]
\centering
\begin{tabular}{|c c c c c c c|}
\hline
Algorithm                & Mean & p25 & Median& p75  & p95  & std  \\ \hline
GLMM                     & 0.68 & 0.68 & 0.69 & 0.71 & 0.72 & 0.05              \\ \hline

Fair GLMM                & 0.67 & 0.68 & 0.69 & 0.70 & 0.71 & 0.05                 \\ \hline
CRLR                     & 0.65 & 0.64 & 0.65 & 0.66 & 0.68 & 0.01                \\ \hline

Fair CRLR                & 0.65 & 0.64 & 0.65 & 0.66 & 0.68 & 0.01                 \\ \hline
LR                       & 0.57 & 0.56 & 0.57 & 0.58 & 0.59 & 0.01                 \\ \hline

Fair LR                  & 0.56 & 0.55 & 0.56 & 0.57 & 0.58 & 0.01                 \\ \hline
\end{tabular}
\caption{Accuracy.}
\label{tabela11}
\end{table}

\begin{table}[ht]
\centering
\begin{tabular}{|c c c c c c c|}
\hline
Algorithm                & Mean & p25 & Median& p75  & p95  & std  \\ \hline
GLMM                     & 0.77 & 0.69 & 0.78 & 0.85 & 0.99 & 0.12              \\ \hline

Fair GLMM                & 0.87 & 0.83 & 0.86 & 0.90 & 0.99 & 0.06                 \\ \hline
CRLR                     & 0.47 & 0.37 & 0.46 & 0.57 & 0.72 & 0.15                \\ \hline

Fair CRLR                & 0.62 & 0.56 & 0.60 & 0.66 & 0.79 & 0.09                 \\ \hline
LR                       & 0.36 & 0.24 & 0.31 & 0.45 & 0.71 & 0.18                 \\ \hline

Fair LR                  & 0.50 & 0.38 & 0.48 & 0.60 & 0.77 & 0.15                 \\ \hline
\end{tabular}
\caption{Disparate impact}
\label{tabela12}
\end{table}

\newpage
\begin{multicols}{2}
\begin{figure}[H]
\centering
\includegraphics[width=6.15cm]{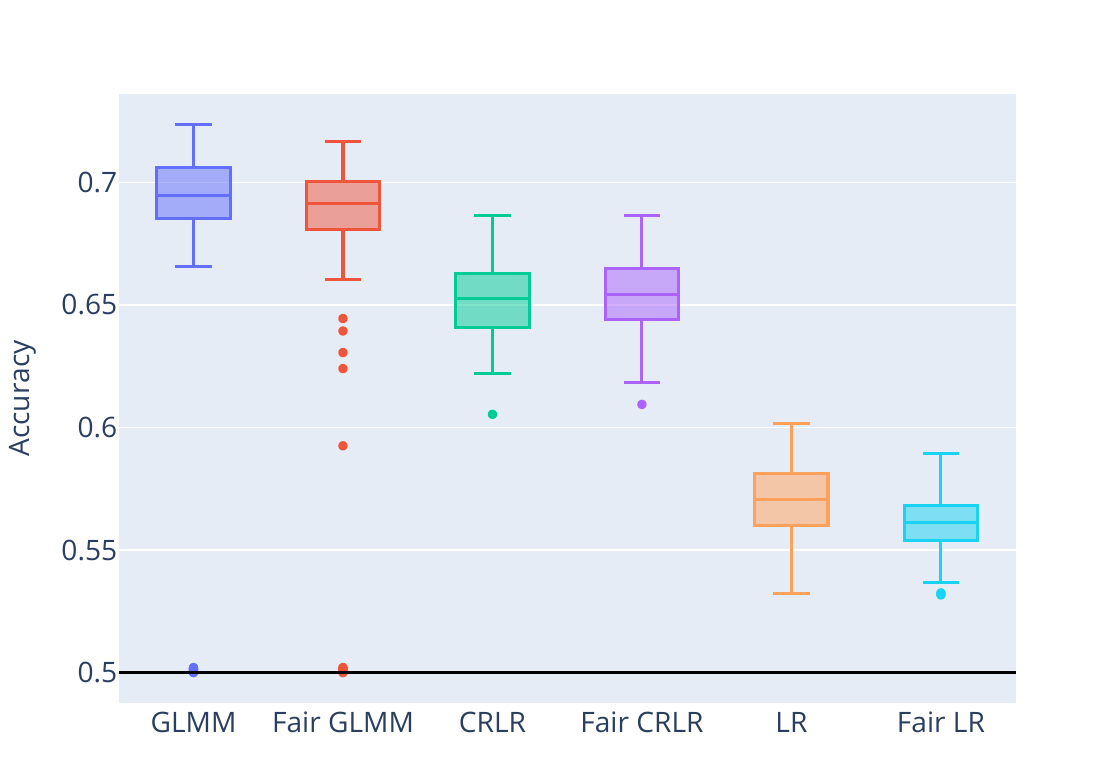}
\caption{Accuracy.}
\label{ACA}
\end{figure}

\begin{figure}[H]
\centering
\includegraphics[width=6.15cm]{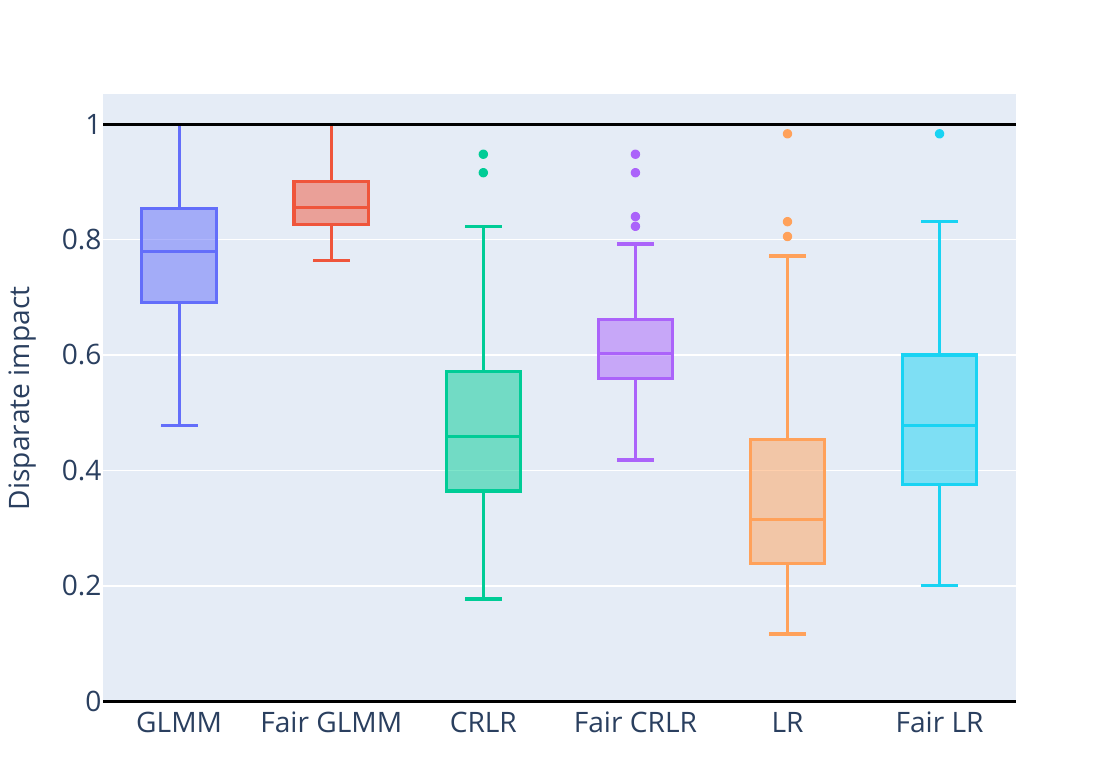}
\caption{Disparate Impact.}
\label{DIA}
\end{figure}
\end{multicols}

We can see that the application is part of scenario \ref{scenario3.1}. Thus, can be seen in Table \ref{tabela11} and in the Figure \ref{ACA} results similar to those previously generated synthetically for the same scenario. That is, we obtained a better accuracy in models that can deal with heterogeneous populations.

Table \ref{tabela12} and Figure \ref{DIA} show that we also obtained an improvement in the disparate impact on the Fair algorithms. Note that make sense since we have a unfair population.


\subsection{Sensitivity analysis}

The Lagrange multiplier strategy, using Karush-Kuhn-Tucker (KKT) conditions, is a powerful tool for analyzing the sensitivity of constrained optimization problems. To perform the sensitivity analysis of the constraints of problems, we use the strategy, described in Appendix \ref{sec:chapter-sensitivity}.

As seen in Section \ref{sec2.2} fairness constraints operate within an interval $[-c, c]$. However, we set $c = 0$. Because this value in the constraint makes the optimization problem as fair as possible, and then, we can say that if we do not obtain a large perturbation in the objective function with $c = 0$, the constraint is useless for the problem. This is because, with other values of $c$, the perturbation would be even smaller.

The sensitivity analysis (SA) or, as it is also known, Lagrange multipliers of fair logistic regression problems is important for understanding how the fairness constraints affect the model's classifications. By analyzing the sensitivity of the model's predictions to changes in the fairness constraints, we can identify which constraints, among the constraints of sensitive features, are most important for achieving fairness. We can also refer to this value as the shadow price, in essence, the shadow price reflects the economic value of relaxing or tightening a constraint in an optimization problem. It represents the marginal impact on the objective function of making a small adjustment to a constraint, that is if the shadow price is $P$, it means that the value of the objective function increase by $P$ if the constraint is relaxed as we can see in \textcite{smith1987shadow}.

In this application, we consider Marital Status, Education, and Housing Loan as sensitive features. The values below refer to the Lagrange multiplier values and the disparate impact values found in the $100$ tests performed on this application. The disparate impact improvement are compared to the regular logistic regression and the accuracy drop are compared with the regular GLMM.

\begin{table}[ht]
\centering
\begin{tabular}{|c c c c|}
\hline
Sensitivity Feature        & SA     & DI improve & AC drop\\ \hline
Housing         & 28.7 & 134.3\%  & 7.5\%    \\ 
Marital Status & 11.9 & 14.4\%  & $<$ 1.0\%    \\ 
Education & 12.5 & 16.6\% & $<$ 1.0\%    \\ 
Marital Status/Education & 13.5/14.7 & 13.9\%/18.5\%  & $<$ 1.0\%    \\  
Housing/Marital Status & 30.9/14.4 & 86.5\%/9.8\%   & 7.9\%   \\ 
Housing/Education & 29.3/13.8 & 82.5\%/16.0\%  & 7.9\%   \\ 
\hline
\end{tabular}
\caption{Sensitivity analysis  table.}
\label{Tabela: Geral}
\end{table}

As we can see in Table \ref{Tabela: Geral}, the sensitive feature constraint that made the biggest difference in the objective function and, therefore, achieved a greater improvement in the Disparate Impact was the housing feature.

We can also note, that in this application simply accounting for the random effect without adding fairness constraints already significantly improves the disparate impact. As expected, adding the fairness constraints, yields even better disparate impact.

%% file: conclusion.tex
\section{Conclusion}\label{sec:conclusion}

In this work, we proposed an algorithm for a fair generalized linear mixed model (GLMM) and a optimization model (CRLR) that allows for controlling the disparate impact of a sensitive feature. This way, a fair classification can be achieved even when the data has an inherent grouping structure. To our knowledge, this has not been proposed before.

We leverage simulations to showcase how our approach overcomes limitations in existing methods. It delivers superior results when group structures significantly impact classification accuracy or fairness.
We also concluded when to use which approach, considering that if we have the information about random effects, the Fair GLMM performs better, otherwise the optimization model is also a good choice.

Furthermore, we applied our method to the Bank marketing dataset. Here, it effectively addressed random effects while mitigating disparate impact associated with the sensitive feature.

Additionally, we explore how KKT conditions can be used to assess feature sensitivity in fair logistic regression.  This analysis helps identify the specific sensitive variable whose disparate impact we aimed to mitigate in the process.

To enhance the framework's capabilities, future work could focus on incorporating additional fairness constraint options within the GLMM. Additionally, efforts to optimize the computational efficiency of the proposed algorithms would be beneficial.

%% file: acknowledgements.tex
\section*{Acknowledgements}

The authors are grateful for the support of the German Federal Ministry of Education and Research (BMBF) for this research project, as well as for the \enquote{OptimAgent Project}.

We would also like to express our sincere appreciation for the generous support provided by the German Research Foundation (DFG) within Research Training Group 2126 \enquote{Algorithmic Optimization}.


%% file: appendix.tex
\section{Sensitivity analysis}
\label{sec:chapter-sensitivity}

The generalization of the sensitivity analysis to multiple sensitive features is straightforward. We simply add a Lagrange multiplier for each sensitive feature. The sensitivity of the objective function can then be determined by the corresponding Lagrange multipliers $\zeta$.  Each $\zeta_k$ is the perturbation of the k-th sensitive feature, with $k = 1,...,K$, where $K$ is the number of sensitive features that we want to use in the tests. And for each sensitive feature, we must use the corresponding $s^k$ and $\bar{s^k}$.

The sensitivity analysis of fair logistic regression problems is important for understanding how the fairness constraints affect the model's classifications. By analyzing the sensitivity of the model's predictions to changes in the fairness constraints, we can identify which constraints are most important for achieving fairness.

The calculation of the sensitivity analysis equations for fair logistic regression problems, fixing $c=0$, is done using the following optimization problem for a fixed $k \in [1, K]$:

\begin{align}
\begin{split}\label{ret11111}
\displaystyle \min_{\beta}\hspace{0.1cm} & -\displaystyle\sum_{\ell=1}^{N} [y_{\ell} \log(m_{\beta}(x^{\ell})) + (1-y_{\ell})\log(1 - m_{\beta}(x^{\ell}))] \\
\textrm{s.t.}\hspace{0.1cm} & \frac{1}{N}\displaystyle\sum_{\ell=1}^{N} \displaystyle (s_{\ell}^k - \bar{s^k})(\beta^\top x^{\ell}) = 0
\end{split}
\end{align}
with $m_{\beta}(x^{\ell})$
\[
m_{\beta}(x^{\ell}) = \dfrac{1}{1 + e^{-(\beta^\top x^{\ell})}}.
\]

Calculating the Lagrangian for problem $\eqref{ret11111}$, we achieve:

\[
\mathcal{L}(\beta, \zeta_k) =  -\sum_{\ell=1}^N \Big[ y_{\ell} \log(m_{\beta}(x^{\ell})) + (1 - y_{\ell})\log(1-m_{\beta}(x^{\ell})) \Big] + \zeta_k\Big(\frac{1}{N} \sum_{\ell=1}^N (s_{\ell}^{k} - \bar{s^k})(\beta^\top x^{\ell}) \Big)
\]

Now, calculating the partial derivative of the Lagrangian with respect to $\beta$ as can be seen in \textcite{hilbe2009logistic}, we have:

\[
\nabla \mathcal{L}(\beta, \zeta_k) = -\sum_{\ell=1}^N \Big( m_{\beta}(x^{\ell}) - y_\ell  \Big)x^{\ell} +  \zeta_k \Big(\sum_{\ell=1}^N (s_{\ell}^{k} - \bar{s^k})(x^{\ell})\Big)
\]

Using the Karush-Kuhn-Tucker (KKT) conditions, as we can see in \textcite{bertsekas2003convex}, we find $\nabla \mathcal{L}(\beta, \zeta_k) = 0$, so:

\begin{equation}\label{eqsa}
    -\sum_{\ell=1}^N \Big( \dfrac{1}{1 + e^{-(\beta^\top x^{\ell})}} - y_\ell  \Big)x^{\ell} + \zeta_k \Big(\sum_{\ell=1}^N (s_{\ell}^{k} - \bar{s^k})(x^{\ell})\Big) = 0
\end{equation}

Since we already have the fixed effects $\beta$'s, the perturbations of all constraints can be obtained by solving the system \eqref{eqsa}. Since the solution found by solving the optimization problem is an optimal point, the KKT conditions are satisfied, so the system has a guaranteed solution.

For the Cluster-Regularized Logistic Regression, fixing $c=0$, the sensitivity analysis is done using the following optimization problem for a fixed $k \in [1, K]$:

\begin{align}
\begin{split}\label{ret1231111}
\displaystyle \min_{\beta,b}\hspace{0.1cm} & -\displaystyle\sum_{i=1}^n \displaystyle\sum_{j=1}^{T_i} [y_{i_j} \log(m_{\beta,b}(x^{i_j})) + (1-y_{i_j})\log(1 - m_{\beta,b}(x^{i_j}))] + \lambda \sum_i^n b_i^2\\
\textrm{s.t.}\hspace{0.1cm} & \frac{1}{N}\displaystyle\sum_{i=1}^n \displaystyle\sum_{j=1}^{T_i} (s_{i_j}^k - \bar{s^k})(\beta^\top x^{i_j} + b_i) = 0
\end{split}
\end{align}
with
\begin{equation*}
    m_{\beta,b}(x^{i_j}) = \dfrac{1}{1 + e^{-(\beta^\top x^{i_j} + b_i)}},
\end{equation*}
calculating the Lagrangian for problem $\eqref{ret1231111}$, we achieve:
\begin{align}
\begin{split}
\mathcal{L}(\beta,b,\zeta_k) &= -\displaystyle\sum_{i=1}^n \displaystyle\sum_{j=1}^{T_i} [y_{i_j} \log(m_{\beta,b}(x^{i_j})) + (1-y_{i_j})\log(1 - m_{\beta,b}(x^{i_j}))] \\ &+ \lambda \sum_{i=1}^n b_i^2 + \zeta_k\Big(\frac{1}{N}\displaystyle\sum_{i=1}^n \displaystyle\sum_{j=1}^{T_i} (s_{i_j}^k - \bar{s^k})(\beta^\top x^{i_j} + b_i) \Big)
\end{split}
\end{align}

Now, computing the partial derivative of the Lagrangian with respect to $\beta$, we have:

\begin{align}
\begin{split}
\nabla_{\beta} \mathcal{L}(\beta,b,\zeta_k) &= -\displaystyle\sum_{i=1}^{n} \displaystyle\sum_{j=1}^{T_i} \Big[y_{i_j} \Big( \frac{x^{i_j}}{1 + e^{(\beta^{\top} x^{i_j} + b_i)}} \Big) - (1- y_{i_j}) \Big( \frac{x^{i_j}}{1 + e^{-(\beta^{\top} x^{i_j} + b_i)}} \Big)\Big] \\ &+ \zeta_k(\frac{1}{N} \displaystyle\sum_{i=1}^{n} \displaystyle\sum_{j=1}^{T_i} (s_{i_j}^k - \bar{s^k})x^{i_j})
\end{split}
\end{align}
and the partial derivative with respect to each $b_i$:
\begin{align}
\begin{split}
\nabla_{b_i} \mathcal{L}(\beta,b,\zeta_k) &= -\displaystyle\sum_{j=1}^{T_i} \Big[y_{i_j} \Big( \frac{1}{1 + e^{(\beta^{\top} x^{i_j} + b_i)}} \Big) - (1- y_{i_j}) \Big( \frac{1}{1 + e^{-(\beta^{\top} x^{i_j} + b_i)}} \Big)\Big] \\ &+ 2\lambda b_i + \zeta_k(\frac{1}{N} \displaystyle\sum_{j=1}^{T_i} (s_{i_j}^k - \bar{s^k}))
\end{split}
\end{align}
that is,
\[
\nabla \mathcal{L}(\beta,b,\zeta_k) = \begin{bmatrix}
\nabla_{\beta} \mathcal{L}(\beta,b,\zeta_k) \\
\nabla_{b} \mathcal{L}(\beta,b,\zeta_k) \\
\end{bmatrix} = \begin{bmatrix}
\nabla_{\beta} \mathcal{L}(\beta,b,\zeta_k) \\
\nabla_{b_1} \mathcal{L}(\beta,b_1,\zeta_k) \\
\vdots \\
\nabla_{b_i} \mathcal{L}(\beta,b_i,\zeta_k) \\
\end{bmatrix}. 
\]


Using the Karush-Kuhn-Tucker (KKT) conditions, we find $\nabla \mathcal{L}(\beta,b,\zeta_k) = 0$ and since we already have the fixed effects $\beta$'s and the random effects $b$'s, the Lagrange multipliers of all constraints can be obtained by solving the system. Since the solution found by solving the optimization problem \eqref{ret12345} - \eqref{ret1333233} is an optimal point, the KKT conditions are satisfied and the system has a guaranteed solution.
